%% file: main.tex
\pdfoutput=1
\documentclass[10pt,twocolumn,letterpaper]{article}

\usepackage[pagenumbers]{cvpr}      %

\input{preamble}
\definecolor{cvprblue}{rgb}{0.21,0.49,0.74}
\usepackage[pagebackref,breaklinks,colorlinks,allcolors=cvprblue]{hyperref}

\title{Body Schema Policy: Body as the Vision--Action Interface\\for Efficient Robot Loco-Manipulation}

\author{Lechen Zhang \quad Chenyuan Chen \quad Muxin Zhou \quad Zewen He \quad Yoshihiko Nakamura\\
Mohamed bin Zayed University of Artificial Intelligence\\
{\normalsize\url{https://www.lechen-zhang.com/BSP/}}}

\begin{document}

\twocolumn[{%
\renewcommand\twocolumn[1][]{#1}%
\vspace*{-12pt}
\maketitle
\vspace{-14pt}
\begin{minipage}{\textwidth}
  \centering
  \includegraphics[width=\textwidth]{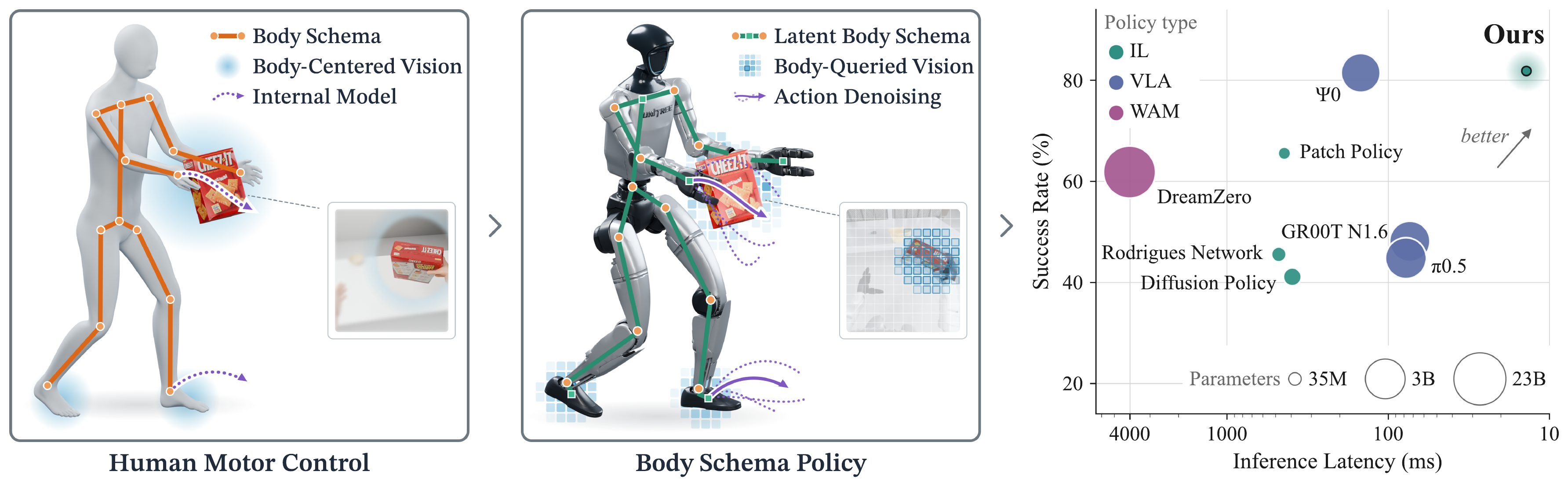}
  \captionof{figure}{\textbf{From the human body schema to a robot's latent body schema.}
  Left: the human body schema links perception with the body and its movement. Middle: \netname{} (\methodshort{}) organizes visual queries and action generation around features on the robot's latent body schema. Right: success rate versus action chunking latency on simulated G1 whole-body tasks. IL: imitation learning; VLA: vision--language--action model; WAM: world-action model.}
  \label{fig:teaser}
\end{minipage}
\vspace{6pt}
}]

\begin{abstract}
Currently, robots can learn whole-body loco-manipulation through scale: the strongest policies have billions of parameters and still need task-specific fine-tuning. Humans have limited processing capacity, yet learn and act efficiently through a body schema that organizes what they see and how they move. We introduce \netname{} (BSP), an imitation learning policy that builds such a schema into the vision--action interface. Its latent body schema places joint and link tokens on the robot's kinematic tree, where learned screw motions carry features between links. Through body-queried vision, each body part retrieves its own visual evidence, which conditions the denoising of its actions. With this inductive bias, BSP returns an action chunk faster than any policy we compare. Compact, body-specific visual readouts also emerge without spatial supervision: manipulation commands depend strongly on task objects, while locomotion commands respond more broadly to the scene. On nine simulated Unitree G1 whole-body tasks, BSP matches the $\Psi_0$ humanoid foundation model with $75\times$ fewer parameters and nearly doubles Diffusion Policy's success. The same holds on a real G1 and a Fetch mobile manipulator.
\end{abstract}

\input{sections/intro}          %
\input{sections/related}        %
\input{sections/method}         %
\input{sections/experiments}    %
\input{sections/conclusion}     %

\section*{LLM Usage Statement}
We used Claude (Anthropic), through Claude Code, and Codex (OpenAI) to edit
and draft manuscript text; to help write training, evaluation, analysis and
plotting code; to help launch and monitor training and evaluation runs; to
help lay out figures; to check the manuscript and its references; and for
literature search. The authors decided the research ideas and the
experimental design, verified all results, checked all AI-assisted text,
code and figures, and take full responsibility for the content of the paper.

{
    \small
    \bibliographystyle{ieeenat_fullname}
    \bibliography{main}
}

\end{document}

%% file: preamble.tex
\usepackage{cuted} %

\usepackage{currfile} %

\usepackage{caption} %

\usepackage{multirow}
\usepackage{adjustbox}   %
\usepackage{pifont}
\usepackage{tikz}      %
\newcommand{\cmark}{\textcolor{green!55!black}{\ding{51}}}   %
\newcommand{\xmark}{\textcolor{red!80!black}{\ding{55}}}   %

\newcommand{\opname}{Neural Screw Operator}      %
\newcommand{\netname}{Body Schema Policy}        %
\newcommand{\methodshort}{\textsc{BSP}}          %
\newcommand{\nsn}{\textsc{BSP}}                  %

%% file: sections/intro.tex
\section{Introduction}
\label{sec:intro}

We study visuomotor policy learning for whole-body loco-manipulation, where a robot must coordinate locomotion and precise manipulation from camera images and proprioception. Recent progress has combined large-scale pretraining with increasingly capable vision--language--action models (VLAs) and world-action models (WAMs)~\citep{pmlr-v305-black25a,bjorck2025gr00t,wei2026psi0,ye2026dreamzero}. These models bring progress from manipulation to loco-manipulation, but with billions of parameters, each must still be fine-tuned on demonstrations for every task before deployment. Humans, in contrast, have limited processing capacity and attention~\citep{marois2005capacity}, yet learn and carry out hard whole-body skills efficiently. This raises our central question: \emph{Can a visuomotor policy learn whole-body skills as efficiently as humans do?}

Human motor control suggests an answer: perception and action are organized around the body. A body schema represents the body's structure, posture, and movement~\citep{head1911sensory,maravita2004tools}. Vision follows the body: visual receptive fields move with the arm~\citep{graziano1994coding}, people look toward task-relevant locations before acting on them~\citep{land2001eye,johansson2001eye}, and focal vision examines detail in the small area where the hands work, while ambient vision guides locomotion through the space at large~\citep{trevarthen1968two}. Movement is refined on the same schema: an internal model predicts how the body will move under a motor command and combines that prediction with sensory feedback to estimate the body's state~\citep{wolpert1995internal}. These observations motivate a robot representation of the body in which an action hypothesis decides what each body part looks at, and what it sees refines the action.

Common visuomotor policy architectures, however, are not designed around such structure: they represent the action as a flat vector of coordinates and the image as features shared by the whole action. For example, Diffusion Policy conditions the action chunk on a pooled visual representation shared across coordinates~\citep{chi2023diffusion}. As a result, how the joints move together, and which part of the image matters to which part of the body, stay hidden from the network and must be learned from demonstrations~\citep{zhang2026rodrigues}. Our forward-kinematics experiments illustrate the first point: generic architectures such as MLPs and Transformers, even with morphology masks, struggle to extrapolate beyond their training range (Section~\ref{sec:exp:fk}). Building this structure into the vision--action interface can make today's policies more efficient.

\begin{figure*}[t]
\centering
\includegraphics[width=\linewidth]{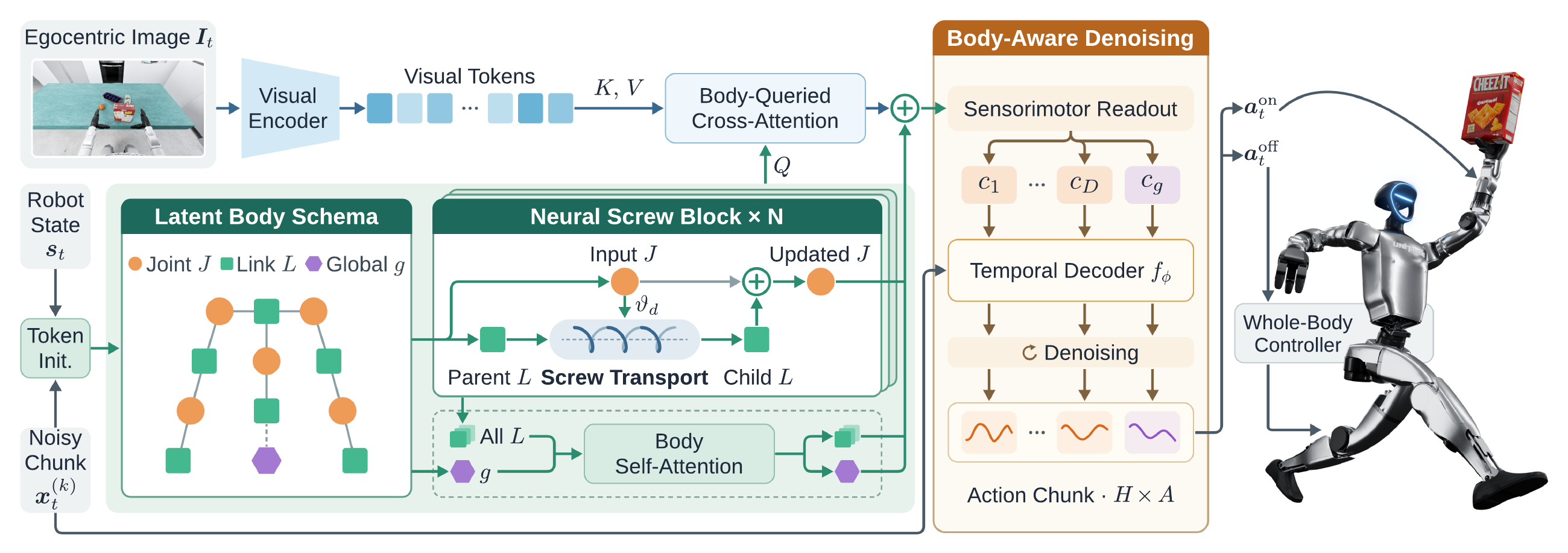}
\caption{\textbf{\netname{} architecture.} The state $s_t$, denoising step $k$, and current action hypothesis $x_t^{(k)}$ initialize joint ($J$), link ($L$), and global ($g$) tokens. Each of the $N$ neural screw blocks transports parent-link features to child links (Eq.~\ref{eq:bilinear}), then updates the corresponding joint tokens. Periodic blocks add body self-attention over links and the global token and body-queried cross-attention, which retrieves image features with body tokens as queries $Q$ and visual tokens as keys and values $K$, $V$; the last such block is drawn. The final schema supplies one condition $c_n$ per action coordinate to a shared temporal decoder $f_\phi$, which predicts action noise.}
\label{fig:arch}
\end{figure*}

To address the problem, we bring the structure of human motor control to robot policies with \netname{} (\methodshort{}), a visuomotor policy for loco-manipulation with three components (Figures~\ref{fig:teaser} and~\ref{fig:arch}). First, a latent body schema places joint and link tokens on the robot's body graph, together with a global token for whole-body features. A \opname{} propagates link features through learned latent screw motions using the graph's parent--child relations. Second, body-queried vision lets body tokens independently retrieve visual evidence from a pretrained encoder's patch tokens. Third, body-aware action denoising uses the resulting schema to condition a shared temporal decoder for each action coordinate. The current noisy action chunk initializes the schema, so its representation and visual queries can change as denoising refines the action.

We evaluate \methodshort{} on two embodiments: a Unitree G1 humanoid, in the SIMPLE benchmark~\citep{wei2026simple} and in the real world, and a wheeled mobile manipulator in ManiSkill-HAB~\citep{shukla2024mshab}. The main findings are:
\begin{enumerate}\setlength{\itemsep}{0pt}
\item \textbf{Action denoising on the latent body schema is fast and accurate.} After five of 100 iterations, the median error of \methodshort{}'s next action on training frames is below 2$^\circ$, against 7--12$^\circ$ for a Diffusion Policy denoiser of the same size, and \methodshort{} returns action chunks faster than any other policy we compare (Section~\ref{sec:exp:why}, Figure~\ref{fig:body_strips}).
\item \textbf{Body-specific visual readouts.} Without spatial supervision, \methodshort{} develops distinct visual sensitivities for manipulation and locomotion: manipulation commands depend strongly on task objects, while base commands respond more broadly to the scene (Section~\ref{sec:exp:why}, Figure~\ref{fig:keyframes}). \methodshort{} also keeps up with foundation models many times its size under distractors, lighting and material changes, and leads every method when poses change (Table~\ref{tab:main}).
\item \textbf{Foundation-model-level task performance from a compact policy.} On simulated G1 whole-body tasks, the 35M-parameter \methodshort{} matches $\Psi_0$'s success rate with $75\times$ fewer parameters and nearly doubles Diffusion Policy's success rate (Figure~\ref{fig:teaser}, Table~\ref{tab:main}). On the real G1 whole-body tasks, it also ties with the $\Psi_0$ foundation model, far ahead of Diffusion Policy (Table~\ref{tab:realmain}). On the Fetch mobile manipulator, it ties with the $\pi_{0.5}$ foundation model and beats both policies of its own size on all three subtasks (Table~\ref{tab:mshab}).
\end{enumerate}

%% file: sections/related.tex
\section{Related work}
\label{sec:related}

\paragraph{Visuomotor policies at scale}
Visuomotor imitation-learning (IL) methods such as Diffusion Policy and ACT learn action chunks from camera observations~\citep{chi2023diffusion,zhao2023act}. Vision--language--action models build on pretrained vision--language representations~\citep{kim2024openvla,black2024pi0,pmlr-v305-black25a,wei2026psi0,bjorck2025gr00t}, while world-action models such as DreamZero jointly model future video and actions~\citep{ye2026dreamzero,nvidia2026cosmos3,yu2026openwam}. These policies solve many manipulation tasks, and the largest have billions of parameters. In loco-manipulation, the same policies command a wheeled base directly or, on humanoids, a large-scale pretrained whole-body controller~\citep{luo2025gr00twbc,luo2026sonic,chen2026holomotion1} that executes their commands while maintaining balance. While scaling data, compute and model size drove most of this progress, we study a complementary axis, efficiency: how explicit body structure can organize perception and action generation.

\paragraph{Body structure in robot policies}
Robot body-schema learning studies representations of a robot's own structure and state~\citep{hoffmann2010body}. Graph policies~\citep{wang2018nervenet,sanchez2018graph}, Body Transformer~\citep{sferrazza2025bot}, MetaMorph~\citep{gupta2022metamorph} and a morphology-aware VLA expert~\citep{suzuki2026morphology} shape their computation by the body's graph, but none models how the joints' motions compose. Implicit kinematic policies~\citep{ganapathi2022implicit}, HDP~\citep{ma2024hdp} and KStar Diffuser~\citep{lv2025kstar} are instead given this composition as exact forward kinematics, which ties them to an idealized robot model. The Rodrigues Network~\citep{zhang2026rodrigues} embeds rotation-based kinematic composition in learned operators, but its published policy takes no images. \methodshort{}, in contrast, propagates latent features with full screw transforms, and its body tokens query the image.

\paragraph{Vision--action interfaces for robot policies}
Visual pretraining has expanded robot perception from compact embeddings~\citep{nair2022r3m,majumdar2023vc1} to dense patch features from vision transformers~\citep{dosovitskiy2021vit,oquab2023dinov2,zhai2023siglip,assran2025vjepa2} and vision--language backbones~\citep{beyer2024paligemma,bai2025qwen3vl}. The vision--action interface determines how these features are selected and used to generate actions. Patch Policy~\citep{zhou2026patchpolicyefficientembodied} lets transformer policies consume pretrained patch tokens directly. Robot-centric Pooling~\citep{zhuang2024rcp} uses proprioception to select image features associated with the robot, while AVA-VLA~\citep{xiao2026avavla} adapts visual attention using instructions and execution history. Instead, \methodshort{} learns this vision--action interface without spatial supervision: joint and global tokens query the image from the latent body schema, and the retrieved evidence conditions their actions.

%% file: sections/method.tex
\section{Method}
\label{sec:method}
\label{sec:network}

Figure~\ref{fig:arch} gives an overview of \netname{}. We begin with the problem and notation (Section~\ref{sec:prob}), followed by the latent body schema (Section~\ref{sec:operator}) and body-queried vision (Section~\ref{sec:network:vtok}). We then describe action denoising (Section~\ref{sec:network:decoder}) and training and inference (Section~\ref{sec:network:diffusion}).

\subsection{Problem definition and notation}
\label{sec:prob}
\label{sec:operator:screw}

\paragraph{Robot visuomotor policy as a denoiser}
From demonstrations, we learn a policy that maps an egocentric image $I_t$ and proprioceptive state $s_t$ to an action chunk $a_t\in\mathbb{R}^{H\times A}$, with $H$ steps and $A$ coordinates. We implement the policy as a denoiser $\epsilon_\phi(x_t^{(k)},k,s_t,V)$, where $V$ denotes the image's patch tokens. At step $k$, the denoiser predicts the noise in the candidate chunk $x_t^{(k)}$, which we call the \emph{action hypothesis}. Each denoising step revises this hypothesis.

\paragraph{Rigid articulated robots for loco-manipulation}
In loco-manipulation tasks, the robot's embodiment and low-level controller determine the policy's command space. We model its articulated body as a rooted kinematic tree with one-DoF revolute, prismatic, or helical joints. The partition $a_t=[a_t^{\mathrm{on}},a_t^{\mathrm{off}}]$ separates on-tree joint targets $a_t^{\mathrm{on}}\in\mathbb{R}^{H\times D}$ from off-tree controller commands $a_t^{\mathrm{off}}\in\mathbb{R}^{H\times(A-D)}$, such as base velocity. The $D$ directly targeted joints and their parent--child relations form the body graph $\mathcal{G}$, a connected subtree with $D+1$ links. Joint $d$, with coordinate $\theta_d$, connects parent link $p(d)$ to child link $c(d)$.

\paragraph{Forward kinematics with screw motions}
Forward kinematics (FK) composes the joint motions along the path from the root to a link~\citep{murray1994mathematical,lynch2017modern}. Poses are $4{\times}4$ homogeneous matrices. For a path with joints $1,\dots,m$, link $\ell$ has pose
\begin{equation}
\label{eq:fkrec}
P_\ell=P_0\,G_1(\theta_1)\cdots G_m(\theta_m)\,M_\ell=\tilde P_\ell\,M_\ell.
\end{equation}
Here $P_0$ is the root pose, $M_\ell$ is the link's zero-configuration pose relative to the root, and each $G_d$ is expressed in that root frame at the zero configuration. The cumulative motion $\tilde P_\ell$ obeys $\tilde P_{c(d)}=\tilde P_{p(d)}\,G_d(\theta_d)$: extending a path appends the next joint motion on the right.
A screw motion combines rotation about an axis with translation along it~\citep{murray1994mathematical}. Each joint motion is generated by a twist $\xi=(\omega,v)\in\mathbb{R}^6$, with angular component $\omega$ and linear component $v$:
\begin{equation}
\label{eq:screw}
G(\theta)=\exp(\theta\hat\xi),\qquad
\hat\xi=\begin{pmatrix}[\omega]_\times&v\\0&0\end{pmatrix}.
\end{equation}
Here $[\omega]_\times$ is the cross-product matrix. The exponential represents revolute, prismatic, and helical motion in one smooth form. For a prismatic joint, $\omega=0$ and $G(\theta)$ is a pure translation by $\theta v$. The closed form of the exponential and its evaluation are given in the supplementary.

\subsection{Latent body schema}
\label{sec:operator}
\label{sec:network:init}

Building on kinematic structure in robot policies~\citep{zhang2026rodrigues}, we define the latent body schema $\mathcal{S}=(J,L,g;\mathcal{G})$. It contains joint tokens $J\in\mathbb{R}^{D\times C_J}$, link tokens $L\in\mathbb{R}^{(D+1)\times C\times4\times4}$, and a global token $g\in\mathbb{R}^{C_g}$. Joint and link tokens condition the on-tree actions; the global token conditions the off-tree commands. Each link has $C$ unrestricted $4{\times}4$ feature channels. Screw transforms act on these channels in the same compositional form that FK uses to propagate poses. Three learned affine maps initialize $J$, $L$, and $g$ from $u=[\operatorname{vec}(x_t^{(k)}),\,e(k),\,s_t]$, where $e(k)$ embeds the denoising step. Every token therefore receives the whole action hypothesis and proprioceptive state, and the schema is rebuilt at each denoising step.

\paragraph{Neural screw block}
\label{sec:network:blocks}
\label{sec:operator:learnable}
Within each denoising step, a stack of neural screw blocks updates the schema. Each block transports parent-link features to their children and updates each joint from its child link. Periodic blocks then apply body self-attention and body-queried visual cross-attention (Section~\ref{sec:network:vtok}). The \opname{} performs the transport. It predicts a latent coordinate $\vartheta_d=w_\theta^\top J_d$ from each joint token, using a head shared across joints (here and below, we omit bias terms). For one channel and one screw transform, the transport is $L_{p(d)}\,G_d(\vartheta_d)$, following FK's right-multiplication structure. The full operator assigns each joint $2R$ learned twists and channel-mixing matrices: $R$ transforms act from the right and $R$ from the left. Writing $p$ and $c$ for the parent and child links and suppressing joint and block indices, the transported feature is
\begin{equation}
\label{eq:bilinear}
\widetilde L_c=\sum_{r=1}^{R}\alpha_r\bigl(L_p\,G_r(\vartheta)\bigr)+\sum_{r=R+1}^{2R}\alpha_r\bigl(G_r(\vartheta)\,L_p\bigr).
\end{equation}
Here $G_r(\vartheta)=\exp(\vartheta\hat\xi_r)$ acts on every feature channel, and $\alpha_r\in\mathbb{R}^{C\times C}$ mixes channels. The block adds the transported features to the link tokens and applies LayerNorm. Each child link then updates the joint that drives it:
\begin{equation}
\label{eq:block}
L\leftarrow\operatorname{LN}\bigl(L+\widetilde L\bigr),\qquad
J_d\leftarrow J_d+W_d\operatorname{vec}(L_{c(d)}).
\end{equation}
At the root, $\widetilde L$ is zero. Periodic body self-attention mixes the link tokens and $g$, leaving the joint tokens unchanged. The unnormalized right-acting transport contains FK: with physical joint coordinates and twists, an identity channel mix, and the root pose as the only nonzero initial feature, a stack as deep as the tree recovers every link's FK product up to a binomial factor (supplementary). In the policy, the body graph fixes connectivity, while the screw parameters and channel mixtures are latent and learned.

\subsection{Body-queried vision}
\label{sec:network:vtok}

The policy accesses the image through body-queried cross-attention. A pretrained DINOv2 encoder~\citep{oquab2023dinov2} maps $I_t$ to patch tokens $V$. Learned positional embeddings $E$ and a projection form the visual memory $\bar V=\operatorname{LN}\bigl((V+E)W_{kv}\bigr)$. At each cross-attention stage, joint $d$ queries this memory using its joint token and parent-link features; the global token supplies a whole-body query.
\begin{equation}
\label{eq:vtok}
\begin{aligned}
J_d&\leftarrow J_d+\operatorname{Attn}_J\bigl([J_d;\operatorname{vec}L_{p(d)}],\,\bar V\bigr),\\
g&\leftarrow g+\operatorname{Attn}_g\bigl(g,\,\bar V\bigr).
\end{aligned}
\end{equation}
Each query updates its own token with visual evidence from the full patch grid. Joint and global queries share the multi-head attention parameters~\citep{vaswani2017attention}, with separate input and output projections. Subsequent blocks distribute this evidence through screw transport and self-attention. At the final cross-attention stage, a joint's visual update directly conditions its own action coordinate. Because queries depend on the current action hypothesis, the retrieved evidence can change during denoising. When multiple cameras are involved, the visual memory holds the patch tokens of all views, each view tagged with the features of the link that carries its camera.

\subsection{Body-aware action denoising}
\label{sec:network:decoder}

The final schema gives one condition per action coordinate:
\begin{equation}
\label{eq:cond}
c_n=\begin{cases}W_c\,[J_n;\operatorname{vec}L_{p(n)}], & n\le D,\\ c_g+b_n, & n>D.\end{cases}
\end{equation}
Here $c_g=W_g\,g$ is the condition read from the global token, and $b_n$ is a learned embedding for each off-tree coordinate. A shared temporal decoder $f_\phi$, built from FiLM-conditioned Conv1D residual blocks, predicts the noise in each column $x_n\in\mathbb{R}^H$ of $x_t^{(k)}$, using its condition $c_n$ and denoising step $k$:
\begin{equation}
\label{eq:decoder}
\hat\epsilon=\epsilon_\phi(x_t^{(k)},k,s_t,V)=\bigl[f_\phi(x_n,c_n,k)\bigr]_{n=1}^{A}.
\end{equation}
Each joint's target is thus denoised from that joint's own part of the schema, and each off-tree command from the global token, with one decoder for all. When a language instruction is given, it is embedded by a frozen text encoder such as CLIP~\citep{radford2021clip} and modulates the joint and global tokens after every block, so that, like the visual evidence, it enters the denoising of every action.

\subsection{Training and inference}
\label{sec:network:diffusion}

We train with the DDPM noise-prediction objective~\citep{ho2020denoising} of Diffusion Policy~\citep{chi2023diffusion}. At a random step $k$, Gaussian noise $\epsilon$ corrupts a demonstrated chunk $a_t$ to form $x_t^{(k)}$, and the network minimizes $\mathcal{L}=\operatorname{MSE}(\epsilon,\hat\epsilon)$. At inference, the policy executes the whole predicted chunk before the next observation. Our serving path caches observation-dependent visual keys and values and replays each denoiser pass as a captured CUDA graph. It produces a 16-step chunk in only 13.9\,ms on an RTX 6000 Ada (supplementary).

%% file: sections/experiments.tex
\begin{figure*}[t]
\centering
\includegraphics[width=\linewidth]{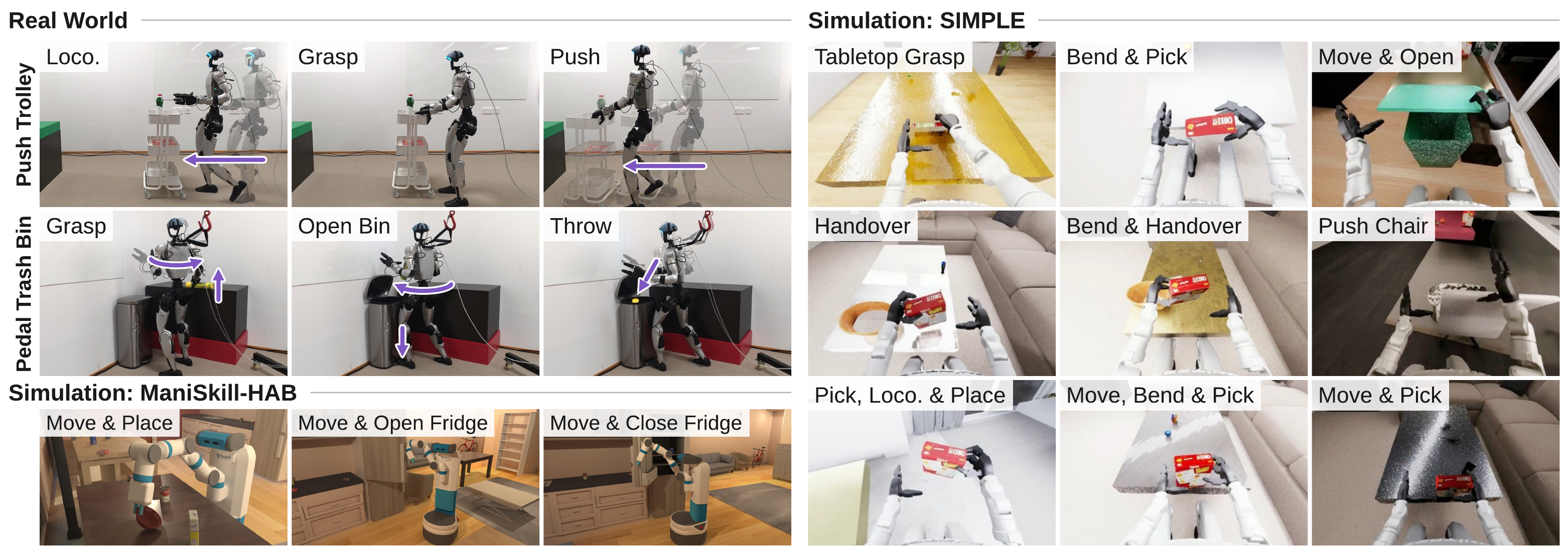}
\caption{Evaluation tasks. Top left: the two real-world G1 tasks, each in three phases.
Purple arrows mark the motion. Bottom left: the three ManiSkill-HAB tasks on the Fetch mobile
manipulator, in third-person views. Right: the nine SIMPLE tasks in egocentric views.}
\label{fig:tasks}
\end{figure*}

\input{sections/tables}

\section{Experiments}
\label{sec:exp}

We compare \nsn{} with state-of-the-art policies on task success and
inference speed, ablate its parts, and analyze its operator, denoising, and
visual representation.

\subsection{Experimental setup}
\label{sec:exp:g1}

\begin{figure*}[t]
\centering
\includegraphics[width=1.0\linewidth]{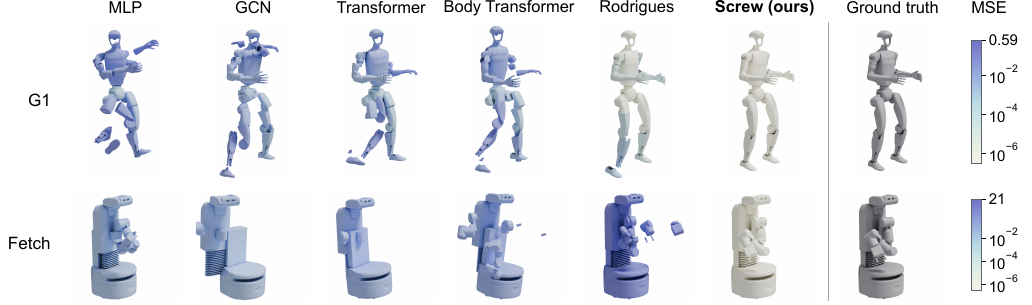}
\caption{Forward kinematics beyond the training range: one test configuration of the G1 (top) and of Fetch (bottom). Each link is drawn displaced by its pose error and colored by that error on a log scale. Fetch is drawn in its base frame, with each model's base translation error removed; the color keeps the full error.}
\label{fig:fk_main}
\end{figure*}

\paragraph{SIMPLE}
We evaluate nine tasks from six categories of the SIMPLE
benchmark~\citep{wei2026simple} (Figure~\ref{fig:tasks}, right), covering
rigid-object grasping (Tabletop Grasp, TTG; Bend \& Pick, BPMP), bimanual handover (Handover, HAND; Bend \& Handover, BHT), mobile
picking (Move \& Pick, XMP; Move, Bend \& Pick, XMBP), whole-body loco-manipulation (Pick, Loco.\ \& Place, LPBT), non-prehensile
pushing (Push Chair, POC), and articulated-object opening (Move \& Open, OTC). Each task has about
100 demonstrations.
Predicted actions are executed by the decoupled whole-body controller
shipped with GR00T N1.6~\citep{luo2025gr00twbc}. Every policy runs in
a closed loop in the benchmark's evaluator on 30 fixed episodes per task,
10 in each of its three evaluation levels: distractors (L0), materials and
lighting (L1), and object and robot poses (L2). Success is the
benchmark's own criterion for each task.

\paragraph{ManiSkill-HAB}
In ManiSkill-HAB~\citep{shukla2024mshab}, we evaluate \nsn{} on a second embodiment: the Fetch mobile manipulator (Figure~\ref{fig:tasks}, bottom left). Its base and torso move by translation, which the screw operator models directly. We train \nsn{}, the Rodrigues Network and Diffusion Policy at a matched size of about 25M trainable parameters, from head and hand RGB images, joint state and a language instruction, on three SetTable subtasks: opening the fridge, closing it, and placing an object at a goal marker that we render in the camera images. We also fine-tune $\pi_{0.5}$ on each subtask. Each model is trained with three seeds, and each seed is evaluated on 1{,}008 episodes.

\paragraph{Real-world G1}
On a real G1, we test two loco-manipulation tasks
(Figure~\ref{fig:tasks}, top left): pushing a trolley and picking up an
item, opening a trash bin with a foot, and throwing the item in. The G1 has 29 body joints and a 7-DoF Dex3-1 hand on each arm; the policies see through its head-mounted RealSense D435i camera. On each task, Diffusion Policy (DP), $\Psi_0$, and \nsn{} learn from the same
15 demonstrations, run on the same RTX 5090 host, and start from the same 10 initial placements
of robot and objects. They command all 43 joints, from legs to fingers, through the
joint-space control policy HoloMotion v1.4~\citep{chen2026holomotion1}.

\subsection{Evaluation results}
\label{sec:exp:results}

\paragraph{Baselines}
We select the main methods of recently published works as baselines. On
the SIMPLE benchmark, they are the IL models DP~\citep{chi2023diffusion},
Patch Policy~\citep{zhou2026patchpolicyefficientembodied}, and the
Rodrigues Network's denoiser~\citep{zhang2026rodrigues}; the VLA models
$\Psi_0$~\citep{wei2026psi0}, $\pi_{0.5}$~\citep{pmlr-v305-black25a}, and
GR00T N1.6~\citep{bjorck2025gr00t}; and the WAM
DreamZero~\citep{ye2026dreamzero}. The supplementary provides all detailed
recipes and statistical conventions.

\paragraph{SIMPLE}
\nsn{} and $\Psi_0$ lead Table~\ref{tab:main}, well ahead of every other method. \nsn{} matches $\Psi_0$
with about $75\times$ fewer parameters and returns an action
chunk more than ten times faster.
$\Psi_0$ comes to each task pretrained on about 800 hours of human video and 30 hours of humanoid data~\citep{wei2026psi0}; \nsn{} learns from the task's demonstrations alone, with a pretrained DINOv2 encoder.
Three tasks, LPBT, XMBP, and BHT, separate the methods: every method except \nsn{} and $\Psi_0$ scores zero
on at least one of them (supplementary). \nsn{} also holds up best when object and robot
poses change: at level L2, where every method scores lowest, it leads every other method (Table~\ref{tab:main}, column L2).
Patch Policy has the same encoder and size as \nsn{} and its own action head; it trails \nsn{} at every level, and most at L2, so neither the encoder nor the size explains \nsn{}'s lead.

\begin{figure*}[t]
\centering
\includegraphics[width=\linewidth]{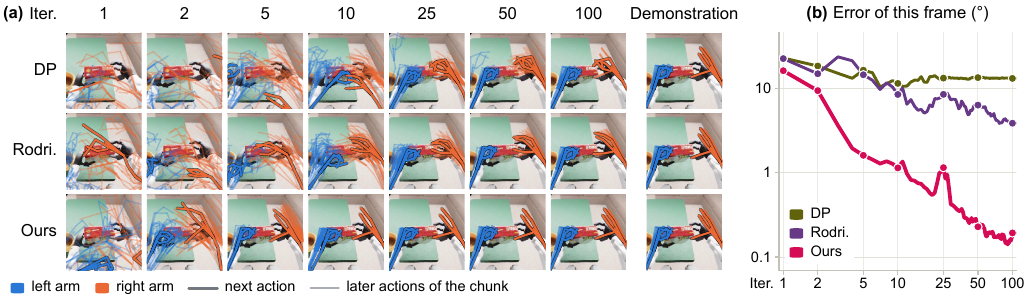}
\caption{Denoising from noise on Handover tasks.
(a) Each row is one denoiser's clean-action estimate after the given number of DDPM
iterations, out of 100, drawn in the head-camera view;
last column: the demonstration. (b) The joint-command error of the next
action at every iteration: the mean absolute difference from the demonstration over the 31
joint commands of the hands, arms, and waist; dots mark the iterations shown in (a). Rodri.: Rodrigues Network.}
\label{fig:body_strips}
\end{figure*}

\paragraph{ManiSkill-HAB}
On Fetch, \nsn{} has the highest success rate among the three policies of matched size, on every subtask and in every seed: 1.6 to 1.8 times that of the better baseline (Table~\ref{tab:mshab}). On the same episodes, $\pi_{0.5}$, with 3.62B parameters, is level with \nsn{} on opening the fridge and behind it on the other two subtasks.

\begin{table}[t]
\caption{ManiSkill-HAB SetTable subtasks with the Fetch mobile manipulator. Success rate (\%), mean $\pm$ standard deviation over three training seeds; 1{,}008 episodes per seed. Params: the whole model as deployed. Policy type: \ptype{ptIL}IL, \ptype{ptVLA}VLA.}
\label{tab:mshab}
\centering
\footnotesize
\setlength{\tabcolsep}{0pt}
\renewcommand{\arraystretch}{1.40}
\begin{tabular*}{\linewidth}{@{}l@{\extracolsep{\fill}} r @{\hspace{0.6em}}ccc@{}}
\toprule
 & Params & Open fridge & Close fridge & Place \\
\midrule
\ptype{ptIL}DP & 47.5M & 18.7 $\pm$ 1.9 & 13.2 $\pm$ 0.3 & 6.5 $\pm$ 0.4 \\
\ptype{ptIL}Rodrigues Network & 47.2M & 18.2 $\pm$ 0.6 & 9.8 $\pm$ 1.9 & 9.3 $\pm$ 2.3 \\
\ptype{ptVLA}$\pi_{0.5}$ & 3.62B & 33.4 $\pm$ 1.2 & 11.4 $\pm$ 0.8 & 13.3 $\pm$ 1.3 \\
\ptype{ptIL}\nsn{} (ours) & 47.4M & \textbf{34.4 $\pm$ 3.6} & \textbf{20.6 $\pm$ 1.6} & \textbf{16.5 $\pm$ 1.6} \\
\bottomrule
\end{tabular*}
\end{table}

\paragraph{Real-world G1}
On the real robot, a trial succeeds when the trolley is pushed in front of the table or when the grasped trash lands in the bin; a fall or a safety stop counts as a failure. \nsn{} and $\Psi_0$ lead again, and Diffusion Policy never completes the trash-bin task (Table~\ref{tab:realmain}).

\begin{table}[t]
\caption{Real-world G1 tasks: successes in 10 trials per task. Params: the whole model as deployed. Time: ms per action chunk on an RTX 5090. Policy type: \ptype{ptIL}IL, \ptype{ptVLA}VLA.}
\label{tab:realmain}
\centering
\footnotesize
\setlength{\tabcolsep}{0pt}
\renewcommand{\arraystretch}{1.40}
\begin{tabular*}{\linewidth}{@{}l@{\extracolsep{\fill}} r r @{\hspace{0.6em}}ccc@{}}
\toprule
 & Params & Time & Push trolley & Pedal trash bin & Total \\
\midrule
\ptype{ptIL}DP & 84.4M & 284 & 3 & 0 & 3 \\
\ptype{ptVLA}$\Psi_0$ & 2.63B & 137 & 7 & 6 & 13 \\
\ptype{ptIL}\nsn{} (ours) & 48.3M & \textbf{17.4} & \textbf{8} & \textbf{8} & \textbf{16} \\
\bottomrule
\end{tabular*}
\end{table}

\subsection{Ablations}
\label{sec:exp:abl}

We ablate \nsn{} on three representative tasks from SIMPLE, with 30 released episodes per task (Table~\ref{tab:abl_main}).

\paragraph{Role of the latent body schema}
We replace the latent body schema with Diffusion Policy's U-Net (last two rows of Table~\ref{tab:abl_main}). With either encoder, the U-Net succeeds less often than the latent body schema (the two rows above). Without the latent body schema, the model largely fails on the two loco-manipulation tasks, LPBT and XMBP, while on the bimanual task HAND, the performance is matched. This shows that the latent body schema matters where locomotion and manipulation must be coordinated. The U-Net here has five times as many parameters because a U-Net of matched size scores only 0 out of 90.

\begin{table}[t]
\caption{Ablation studies. We study the effects of the latent body schema, body-queried vision, and the DINOv2 encoder on 3 tasks in SIMPLE; successes out of 30 per task, one seed each. \xmark{} replaces them, in column order, with DP's U-Net, one pooled vector, and a ResNet-18 trained from scratch.}
\label{tab:abl_main}
\centering
\footnotesize
\setlength{\tabcolsep}{3.4pt}
\newcommand{\hdr}[1]{{\renewcommand{\arraystretch}{0.9}\begin{tabular}[c]{@{}c@{}}#1\end{tabular}}}
\renewcommand{\arraystretch}{1.40}
\begin{tabular}{@{}ccc cccc@{}}
\toprule
\hdr{Latent body\\schema} & \hdr{Body-queried\\vision} & DINOv2 & \textbf{LPBT} & \textbf{XMBP} & \textbf{HAND} & Total \\
\cmidrule(r){1-3}\cmidrule(l){4-7}
\cmark & \cmark & \cmark & 23 & 28 & 24 & \textbf{75} \\
\cmark & \xmark & \cmark & 18 & 24 & 18 & 60 \\
\cmark & \xmark & \xmark & 10 & 8 & 13 & 31 \\
\xmark & \xmark & \cmark & 5 & 0 & 19 & 24 \\
\xmark & \xmark & \xmark & 6 & 0 & 14 & 20 \\
\bottomrule
\end{tabular}
\end{table}

\paragraph{Body-queried vision versus a pooled vector}
Letting each body part query the patch tokens, in place of one pooled DINOv2 vector, improves all three tasks. Most of the gain comes at L2, where poses change: the pooled model drops there, while the full model holds its level (supplementary). Section~\ref{sec:exp:why} analyzes what each part reads.

\paragraph{Role of the DINOv2 encoder}
With one pooled vector as the image input, replacing DINOv2 with a ResNet-18 trained from scratch halves the total: \nsn{} benefits largely from strong DINOv2 features. The reliance is mutual: without the latent body schema, the policy gains little from DINOv2. The latent body schema thus unlocks the policy's potential to benefit from state-of-the-art visual encoders.

\begin{figure*}[t]
\centering
{\offinterlineskip\includegraphics[width=\linewidth,trim=0 92.54bp 0 0,clip]{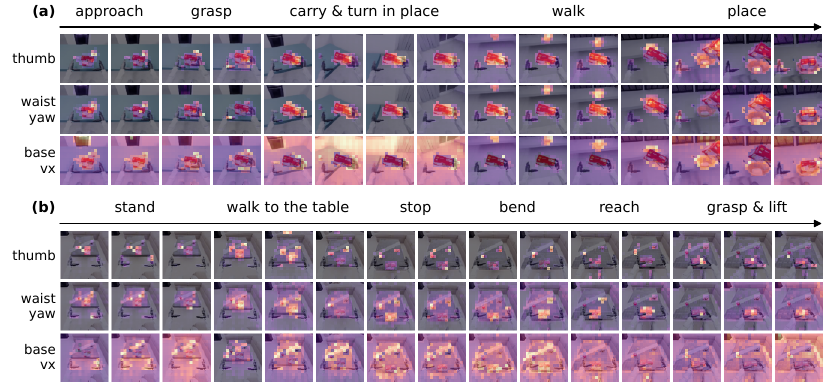}\par\vspace{12pt}%
\includegraphics[width=\linewidth,trim=0 0 0 92.5bp,clip]{figures/fig_keyframes_ab_rows.pdf}\par}
\caption{What three parts read over one LPBT episode (a) and one XMBP
episode (b). Sensitivity
of \nsn{}'s right-thumb, waist-yaw and base-velocity commands to the
patch tokens the denoiser reads, at 15 frames of each
episode; each heat map is scaled to its own maximum.}
\label{fig:keyframes}
\end{figure*}

\subsection{Analysis}
\label{sec:exp:why}

\paragraph{Operator expressivity}
\label{sec:exp:fk}
We first test the operator's expressivity alone, on forward kinematics beyond the training range. Each model maps a root pose and joint configuration to every link's metric pose; it trains on the middle half of each joint's range and is tested on all of it (ten robots, five seeds; supplementary). The error is the mean squared error (MSE) between predicted and true link poses, at metric scale. Generic networks fit inside the training range and fail beyond it: on every robot, the MLP, Transformer, Body Transformer and graph networks average at least $11\times$ our error, and their predicted bodies come apart (Figure~\ref{fig:fk_main}). Against the rotation-only Rodrigues operator, the screw operator has the lower error on the G1 and on all four robots with prismatic joints; on Fetch, whose base translates, the gap is seven orders of magnitude. On the five other robots, which have only revolute joints, the Rodrigues operator is ahead by 2--6$\times$, while ours is the only network with error below $10^{-4}$ on all ten robots.

\paragraph{Denoising efficiency}
Conditioned on the latent body schema, \nsn{}'s action estimate comes close to
the demonstration within the first few denoising iterations. Figure~\ref{fig:body_strips} shows this on one Handover frame, next to DP and a Rodrigues Network of the same size that read the same patch tokens. The gap is not particular to this frame: over 256 training frames per task, \nsn{}'s median error after five iterations is below 2$^\circ$ on each of three tasks, against 7--12$^\circ$ for DP (supplementary). This pays off at deployment: ten DDPM steps are sufficient for the results in
Table~\ref{tab:main}, where \nsn{} is also the fastest policy we compare.

\paragraph{Visual representation compactness}
Each joint develops a task-specific visual readout without any spatial supervision.
We measure how strongly each command depends on each image patch: the gradient of the command with respect to the patch tokens at the last denoising step (Figure~\ref{fig:keyframes}; supplementary). The map is specific to the body part. On LPBT, \nsn{}'s arm, hand and waist joints put 31--46\% of their sensitivity on the box, which covers 10--15\% of the image, while its base commands read the whole scene as the robot turns and sets off with the box; on XMBP, the thumb command reads a few patches and the base command most of the scene. In a U-Net that cross-attends to the same tokens, every command reads nearly the same map. The policy also acts on what each part reads (supplementary). Blanking the tenth of the image that an upper-body part is most sensitive to moves its commands 3.7--7.0$\times$ as far as blanking as many random patches, and further than blanking the base's tenth. The base commands move under either blanking: the base uses the whole scene. The latent body schema thus sets both how each part moves and what it looks at. The division it arrives at, detail in a small area for the hands and the space at large for locomotion, is that of focal and ambient vision in primates (Section~\ref{sec:intro})~\citep{trevarthen1968two}. We built in the body; the division of vision emerged.

%% file: sections/tables.tex
\definecolor{ptIL}{HTML}{2F8F82}
\definecolor{ptVLA}{HTML}{5D6FA6}
\definecolor{ptWAM}{HTML}{A4588F}
\DeclareRobustCommand{\ptype}[1]{\tikz[baseline=0.02ex]{\fill[#1,rounded corners=0.3ex] (0,0) rectangle (1.25ex,1.25ex);}\hspace{0.45em}}

\begin{table}[t]
\caption{Nine representative whole-body humanoid tasks on \mbox{SIMPLE}: successes on each of the three evaluation levels (L0/L1/L2), 10 fixed episodes per task per level. Params: the whole model as deployed. Time: ms per action chunk on an RTX 6000 Ada. Policy type: \ptype{ptIL}IL, \ptype{ptVLA}VLA, \ptype{ptWAM}WAM. \textbf{Bold}\,=\,best, \underline{underline}\,=\,second best; tied totals share a mark.}
\label{tab:main}
\centering
\footnotesize
\setlength{\tabcolsep}{0pt}
\renewcommand{\arraystretch}{1.40}
\begin{tabular*}{\linewidth}{@{}l@{\extracolsep{\fill}} r r @{\hspace{0.9em}}c c c c@{}}
\toprule
Method & Params & Time & L0 & L1 & L2 & Total \\
\midrule
\ptype{ptIL}DP~\citep{chi2023diffusion} & 84.3M & 394.0 & 43 & 42 & 26 & 111 \\
\ptype{ptIL}Rodrigues Network~\citep{zhang2026rodrigues} & 61.5M & 477.8 & 41 & 45 & 37 & 123 \\
\ptype{ptIL}Patch Policy~\citep{zhou2026patchpolicyefficientembodied} & 34.9M & 441.4 & 65 & \underline{66} & 46 & 177 \\
\ptype{ptVLA}$\pi_{0.5}$~\citep{pmlr-v305-black25a} & 3.62B & 77.7 & 43 & 40 & 38 & 121 \\
\ptype{ptVLA}GR00T N1.6~\citep{bjorck2025gr00t} & 3.29B & \underline{73.5} & 48 & 43 & 39 & 130 \\
\ptype{ptVLA}$\Psi_0$~\citep{wei2026psi0} & 2.63B & 148.2 & \textbf{81} & \textbf{76} & \underline{63} & \underline{220} \\
\ptype{ptWAM}DreamZero~\citep{ye2026dreamzero} & 22.9B & 4026.5 & 64 & 58 & 45 & 167 \\
\midrule
\ptype{ptIL}\nsn{} (ours) & 35.0M & \textbf{13.9} & \underline{74} & \textbf{76} & \textbf{71} & \textbf{221} \\
\bottomrule
\end{tabular*}
\end{table}

%% file: sections/conclusion.tex
\section{Conclusion}
\label{sec:conclusion}

We introduced \netname{}, a human-inspired visuomotor policy that coordinates visual perception and action generation through a latent body schema. Its \opname{} carries features along the robot's kinematic tree with learned screw motions, and each body part retrieves its own visual evidence through the schema. With the schema, the denoiser reaches the action within the first few iterations: \methodshort{} is the fastest policy we compare, matches the $\Psi_0$ foundation model with $75\times$ fewer parameters, and nearly doubles Diffusion Policy's success on nine G1 whole-body tasks. On the Fetch mobile manipulator, it leads policies of matched size on all three subtasks, and on a real G1 it ties with $\Psi_0$. This gives one answer to our opening question: organized around its own body, a visuomotor policy learns whole-body skills with a small fraction of the parameters, and none of the robot pretraining that foundation models use.

\section{Limitations and future work}
\label{sec:limitations}
Due to limited compute, we could not scale the training and evaluation on SIMPLE to more seeds. \methodshort{} relies on the kinematic meaning of joint commands. Whole-body controllers that are commanded in a latent space, such as SONIC~\citep{luo2026sonic} and BFM-Zero~\citep{li2026bfmzero}, replace the joint commands with latent coordinates that belong to no body part, and the latent body schema is less effective there than in joint space (supplementary). Natural next steps are: (1) cross-embodiment policies that condition on the latent body schema, (2) integration into other robot learning policies such as WAMs and VLAs, and (3) compact libraries of atomic skills for agentic systems: a hundred \methodshort{} policies have fewer parameters than one $\pi_{0.5}$.